\documentclass[10pt,journal,compsoc]{IEEEtran}
\ifCLASSOPTIONcompsoc
  \usepackage[nocompress]{cite}
\else
  \usepackage{cite}
\fi

\usepackage{graphicx}
\usepackage{subfigure}
\usepackage{amsmath}
\usepackage{multirow}
\usepackage{booktabs}
\usepackage{amssymb}

\ifCLASSINFOpdf
\else
\fi
\begin{document}
%
\title{Incomplete Multi-view Clustering via Diffusion Completion}
%
%
%
%

\author{Sifan~Fang
}

\IEEEtitleabstractindextext{%
\begin{abstract}
  Incomplete multi-view clustering is a challenging and non-trivial task to provide effective data analysis for large amounts of unlabeled data in the real world. All incomplete multi-view clustering methods need to address the problem of how to reduce the impact of missing views. To address this issue, we propose diffusion completion to recover the missing views integrated into an incomplete multi-view clustering framework. Based on the observable views information, the diffusion model is used to recover the missing views, and then the consistency information of the multi-view data is learned by contrastive learning to improve the performance of multi-view clustering. To the best of our knowledge, this may be the first work to incorporate diffusion models into an incomplete multi-view clustering framework. Experimental results show that the proposed method performs well in recovering the missing views while achieving superior clustering performance compared to state-of-the-art methods.
\end{abstract}

}

\maketitle

\IEEEdisplaynontitleabstractindextext

%
\IEEEpeerreviewmaketitle

\IEEEraisesectionheading{\section{Introduction}\label{sec:introduction}}

%
%
%
%

 

\begin{figure*}
  \includegraphics[width=\textwidth]{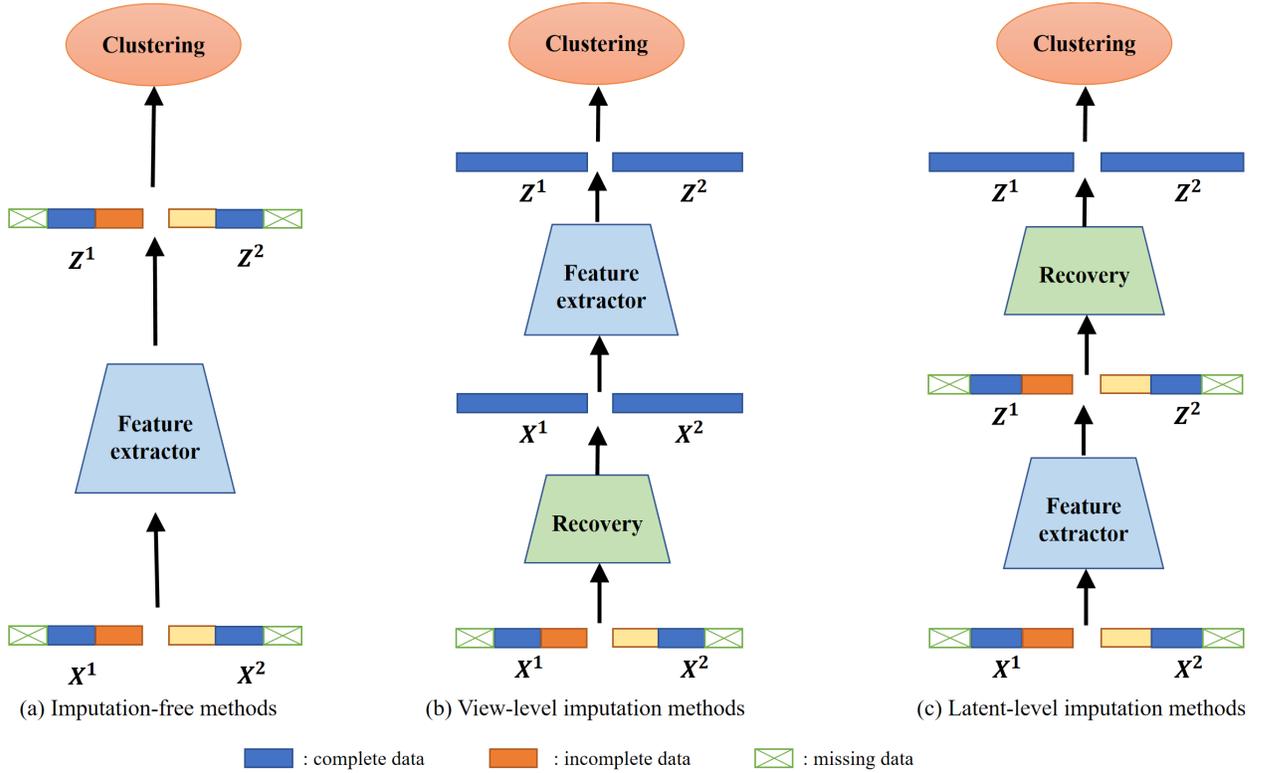}
  \caption{Schematic diagram of the classification of IMVC methods. We categorize IMVC methods into three categories based on their treatment of the missing views. (a) imputation methods, which do not restore missing views and design a machine learning algorithm to implement incomplete multi-view clustering; (b) View-level methods restore the missing views in the original view space and then use existing MVC clustering methods or self-designed MVC methods to achieve multi-view clustering; (c) Latent-level imputation methods, which differ from (b) in restoring missing views in latent space.}
  \label{fig:method_types}
\end{figure*}

\IEEEPARstart{M}{ulti-view} data are widely present in the real world, which refer to data composed of different modalities, views, or forms. In the case of security identification, users can use face recognition, fingerprint recognition, or password recognition to pass security identification. And, there are some more specialized and rigorous security identification systems that require three identifications to be completed at the same time to pass security. Compared with single-view data, multi-view data requires more labor cost to label, so many multi-view unsupervised learning methods \cite{Xu_2021_ICCV, NEURIPS2021_10c66082, Xu_2022_CVPR, wang2019gmc, liu2021one, peng2019comic, chen2020multi} have been developed, namely multi-view clustering (MVC).  
In practice, several views of multi-view data are often missing due to various unavoidable factors. However, the above MVC methods are based on the assumption that the multi-view data is complete, which fails to meet the needs of real-world scenarios. For example, in the case of health care, some patients may not be able to test certain items due to their specific causes or equipment failures in hospitals. Many Incomplete Multi-view clustering methods (IMVC) have been developed \cite{lin2021completer, lin2022dual, ijcai2020p447, zhang2021one, wen2021unified, xu2022deep, tang2022deep, wang2021generative} to solve this situation. 
All IMVC approaches face two challenges, namely, how to handle the missing views and how to guide the clustering of multi-view data. According to these methods, we classify them into three categories, imputation-free, view-level imputation, and latent-level imputation, as shown in Figure \ref{fig:method_types}. The first imputation-free methods \cite{xu2022deep, li2014partial, wen2021structural} do not recover the missing views and use specific machine learning algorithms to reduce the impact of the missing views. However, these methods do not fully exploit multi-view data's consistency and complementary information. The second view-level imputation methods \cite{tang2022deep, wang2021generative, xu2019adversarial} and the third latent-level imputation methods \cite{lin2021completer, zhao2016incomplete, zhang2021one} are both imputation methods. The difference between them is the space to recover the missing view. One is the original view space, and the other is the latent space of views. Both approaches face the same problem of how to recover the missing views.  

It has been observed that some IMVC methods \cite{wang2021generative, wang2022adversarial, xu2019adversarial} use adversarial generation networks as the missing views recovery method, that is, the existing generation methods can be used as the missing views recovery methods. In recent years, diffusion models \cite{sohl2015deep, ho2020denoising, song2019generative} have achieved success in many fields of generation, such as image synthesis \cite{rombach2022high, dhariwal2021diffusion}, audio synthesis \cite{kong2020diffwave, chen2020wavegrad}, text-to-image translation \cite{ramesh2022hierarchical, saharia2022photorealistic}, image super-resolution \cite{li2022srdiff, saharia2022image}, etc. Based on what we have observed and the dynamic development of the diffusion models, we propose the method of Incomplete Multi-view Clustering via Diffusion Completion (IMVCDC).

Our designed IMVCDC belongs to the third IMVC method, the latent-level imputation method, which has two advantages. First, it recovers missing views and takes full advantage of the complementarity of the multi-view data, which facilitates subsequent clustering. The second is to complete the missing views in the latent space, which can significantly reduce the view completion cost. The framework of IMVCDC is straightforward and consists of three modules: view reconstruction, diffusion completion, and contrastive clustering. View reconstruction can be used to obtain efficient low-dimensional representations through auto-encoder reconstruction. Then, the incomplete multi-view data is transformed into the complete multi-view data by diffusion models. Finally, we use contrastive learning to obtain clustering categories for multi-view data. Experimental results show that our proposed IMVCDC performs better than the state-of-the-art IMVC methods, which validates the effectiveness of IMVCDC.

\section{Related work}
This section, briefly presents developments in the field relevant to our work, namely incomplete multi-view clustering and diffusion models.

\subsection{Incomplete Multi-view Clustering}
The existing incomplete multi-view clustering methods can be divided into imputation-free methods \cite{xu2022deep, li2014partial, wen2021structural}, view-level imputation methods \cite{tang2022deep, wang2021generative, xu2019adversarial}, and latent-level imputation methods \cite{lin2021completer, zhang2021one, zhao2016incomplete}. This work \cite{xu2022deep} proposes an imputation-free and fusion-free IMVC framework that mines the complementarity of multi-view by mapping the embedding features of the complete multi-view data into high-dimensional space. PVC \cite{li2014partial} establishes a latent subspace that brings different views of the same sample close to each other and separates similar samples of the same view from each other. SDIMC-net \cite{wen2021structural} simultaneously explores the high-level features and high-order geometric information and uses the weighted fusion approach to reduce the influence of the missing views to obtain a consistent expression. This work \cite{xu2019adversarial} searches for the common latent space of multi-view data and uses adversarial generation networks to recover the missing views, obtaining clustering structures through an aligned clustering loss. This work \cite{tang2022deep} proposed a new framework, that is, to dynamically complete missing views from learned semantic neighbors and automatically select imputed samples for training. This framework improves the performance and safety of multi-view clustering. In \cite{wang2021generative}, multi-view encoders are trained to learn the common low-dimensional representation and to use the cycle-consistent generative adversarial networks to recover the missing views under the condition of shared representation. These two steps facilitate each other to learn a better multi-view clustering structure. In this work \cite{lin2021completer}, the mutual information between different views is maximized by contrast learning, and the missing views are restored by dual prediction based on minimizing the conditional entropy of different views, thus improving the consistent learning of multi-view and missing views completion. This work \cite{zhao2016incomplete} obtains a compact global structure by using the Laplace term to map incomplete multi-modal data into a complete representation in a common subspace. In this work \cite{zhang2021one}, the imputation of multi-view completion and clustering tasks are integrated into a single optimization process so that the learned consensus matrix can directly help the final clustering task.

\begin{figure*}
  \includegraphics[width=\textwidth]{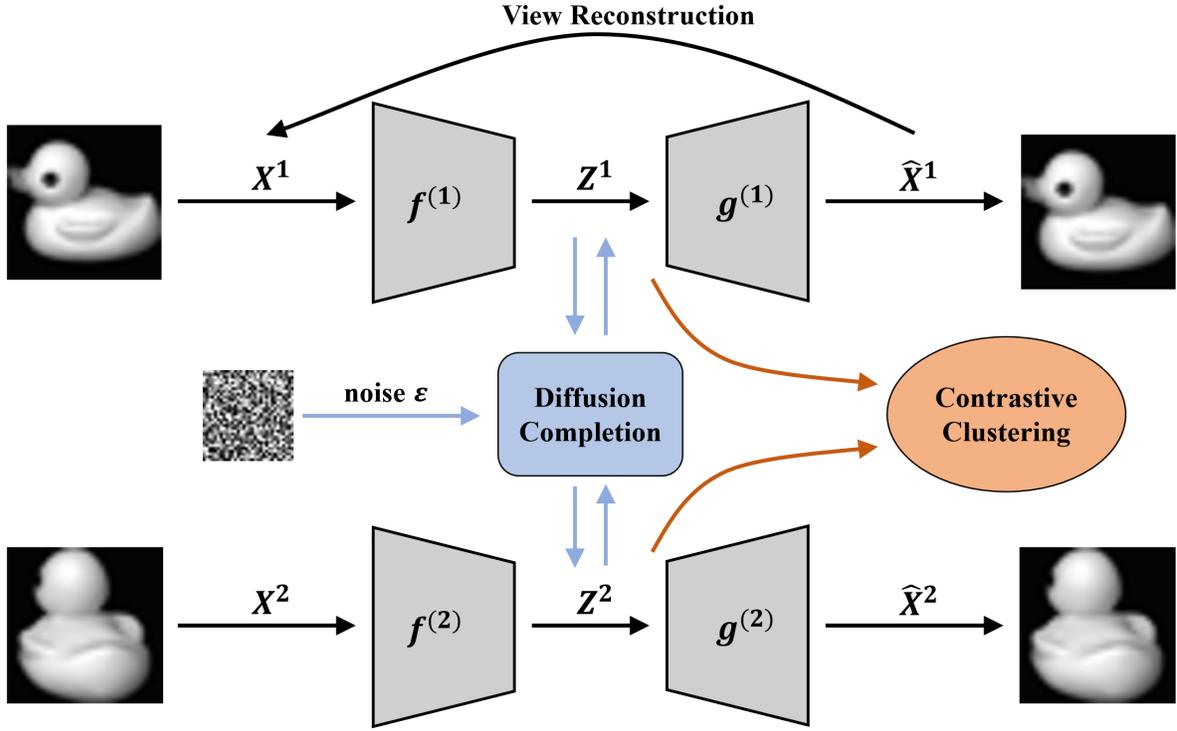}
  \caption{Overview of IMVCDC. The two-view data is used for the presentation. Our approach consists of three parts: perspective reconstruction, diffusion completion, and contrastive clustering. We construct an autoencoder for each view to obtain the latent representation by reconstructing the view. Diffusion completion is conditioned on the latent representation $ Z^1(Z^2) $ of the existing view, and the input Gaussian noise solution constitutes the latent representation $ Z^2(Z^1) $ of the missing view. Contrastive clustering maximizes the similarity between the same sample perspectives and minimizes the similarity between different sample perspectives to obtain the class information of the samples and complete the clustering.}
  \label{fig:overview}
\end{figure*}

\subsection{Diffusion Models}
Generative adversarial networks have succeeded in incomplete multi-view clustering \cite{wang2021generative, wang2022adversarial, xu2019adversarial}, but GANs are faced with the problem of training instability and mode collapse \cite{brock2018large, liu2019spectral}. Derived from the principle of dynamic thermodynamics, diffusion models \cite{sohl2015deep} have the advantages of stable training and diversity of generated results compared with GANs. Diffusion models define a forward diffusion process, where noise is gradually added to the data, and a reverse diffusion process is learned to guide the desired data generation from the noise. With the rapid development of diffusion models, it has outperformed GANs in many fields, such as image generation \cite{ramesh2022hierarchical, saharia2022photorealistic}. In order to make the diffusion models generate data with specific classes, this work \cite{dhariwal2021diffusion} trained a classifier to guide the diffusion process toward conditional information, and subsequently developed many conditional-guided diffusion models \cite{ho2022classifier, nichol2021glide}.
Inspired by the above, we propose IMVCDC, which is motivated by the idea of using existing view information to recover the missing views via diffusion models.

\section{Methodology}
In this section, we propose a new incomplete multi-view clustering method called Incomplete Multi-view Clustering via Diffusion Completion (IMVCDC). As illustrated in Figure \ref{fig:overview}, IMVCDC consists of three modules: view reconstruction, diffusion completion, and contrastive clustering.

\subsection{View Reconstruction}

Given a data set $ X={X^1,X^2} $, with a total of $ n $ samples, and $ X^v_i $ is used to represent the $v$-th view of the $i$-th sample. The index matrix $M \in \mathcal{R}^{n\times2}$ is introduced to represent the missing views:
\begin{equation}
    M^v_i = \left\{
    \begin{array}{cl}
    1 & \mbox{the $v$-th view of the $i$-th sample is observable}\\
    0 & \mbox{otherwise}\\
    \end{array} \right.
  \label{eq:index_matrix}
\end{equation}

Each sample has at least one view, and the missing rate of incomplete multi-view data is denoted by $ \eta = \tilde{n}/n $, where $\tilde{n}$ denotes the number of incomplete multi-view data.

Autoencoders are widely used in the field of multi-view clustering \cite{Xu_2021_ICCV, Xu_2022_CVPR}, often used to extract the latent representation of multi-view data. We construct an autoencoder for each view separately and obtain the latent representation of the view by reconstructing it. The objective function of the reconstruction is as follows:
\begin{equation}
    L_{rec} = \sum_{v=1}^{2} \sum_{i=1}^{n} M_i^v \rVert X_i^v-g^{(v)}(f^{(v)}(X_i^v)) \lVert _2^2
    \label{eq:l_rec}
\end{equation}

Here, $g^{(v)}$ and $f^{(v)}$ are the encoder and decoder for the $v$-th view. Thus, the latent representation of the $v$-th view of the $i$-th sample is as follows:
\begin{equation}
    Z_i^v = f^{(v)}(X_i^v)
    \label{eq:z}
\end{equation}

\subsection{Diffusion Completion}
Many IMVC \cite{lin2021completer, lin2022dual, zhang2021one} methods complete the missing views based on the learned latent space. The most significant difference between us and these methods is that we use the diffusion models to complete the missing views. Diffusion models \cite{sohl2015deep, ho2020denoising} are probabilistic models that learn the real distribution of data by gradually de-noising normal distribution variables. Its denoising process is the reverse process of a fixed Markov chain of time length T, and the corresponding optimization objective is as follows:
\begin{equation}
    L_{dm} = \mathbb{E}_{x,\epsilon~N(0,1)}[\lVert \epsilon-\epsilon_{\theta}(x_t,t) \rVert_2^2]
\end{equation}

Where $t$ is uniform from $\{1,\dots,T\}$ sampling, $\epsilon$ is the noise sampled randomly from the normal distribution with mean 0 and variance 1, $x_t$ is the noise version at time $t$ of input data $x$, $\epsilon_{\theta}$ is the denoising model to be learned.

We hope that the results generated by the diffusion model are the missing view data, and thus a conditional mechanism is introduced to guide the diffusion model to generate the desired results accurately. Conditional mechanisms are widely used in diffusion models, such as image-to-image translation \cite{saharia2022palette}, text-to-image translation \cite{ramesh2022hierarchical,ramesh2022hierarchical}, and image synthesis \cite{ho2022classifier, rombach2022high, pinaya2022brain}. As an example of two-view data, the second view is used as a condition to guide the diffusion model to generate the missing first-view data. The conditional $L_{dm}$ can be formulated as:
\begin{equation}
    L_{dm} = \mathbb{E}_{X^1,\epsilon~N(0,1)}[\lVert \epsilon-\epsilon_{\theta}(X_t^1,t,X^2) \rVert_2^2]
\end{equation}

Compared with diffusion in the original view space, efficient and low-dimensional latent space is more suitable for diffusion models \cite{ramesh2022hierarchical, rombach2022high}. This approach has the advantages of (i) focusing on important semantic information of data, (ii) greatly reducing the amount of computation and improving computational efficiency, and (iii) being conducive to subsequent clustering. Finally, the objective function of diffusion completion can be written
\begin{equation}
    L_{dm} = \mathbb{E}_{Z^1,\epsilon~N(0,1)}[\lVert \epsilon-\epsilon_{\theta}(Z_t^1,t,Z^2) \rVert_2^2]
    \label{eq:l_dm}
\end{equation}

In the denoising model $\epsilon_{\theta}$, we use the attention mechanism \cite{vaswani2017attention} as a conditional mechanism to interact information between views and guide diffusion completion to generate the corresponding missing view data.
\begin{equation}
    \begin{array}{c}
    Attention(Q,K,V) = softmax(\frac{QK^T}{\sqrt{d}})\cdot V\\
    \\
    Q=W_Q \cdot \varphi(Z_t^1),K=W_K \cdot \tau(C^2),V=W_V \cdot \tau(C^2)\\
    \end{array}
\end{equation}

Here, $\varphi(\cdot)$ and $\tau(\cdot)$ are both mapping functions and $W_Q$, $W_K$, and $W_V$ are learnable mapping matrices.

\begin{table}
  \centering
  \begin{tabular}{@{}lc@{}}
    \toprule
    {\bf Algorithm 1} Incomplete Multi-view Clustering via Diffusion Completion \\
    
    \midrule
    {\bf Input:} Dataset $X={X^1,X^2}$, index matrix $M \in \mathcal{R}^{n\times2}$, number of \\
    \qquad clusters $k$, max iterations $T_{rec}$, $T_{dm}$, and $T_{clu}$ \\
    
    {\bf Output:} Cluster prediction $\bar{Y}$ Initialize the parameters of the encoder $f$, \\ \qquad decoder $g$, denoising model $\epsilon_{\theta}$, and all MLP layers of contrastive \\ \qquad clustering \\
    
    {\bf Step 1:} Train encoder $f$ and decoder$g$ \\
    {\bf for} $t=0$ to $T_{rec}-1$ {\bf do} \\
    \qquad Update encoder $f$ and decoder$g$ with all data by Equation \ref{eq:l_rec} \\
    {\bf end for} \\

    {\bf Step 2:} Train denoising model $\epsilon_{\theta}$ \\
    {\bf for} $t=0$ to $T_{dm}-1$ {\bf do} \\
    \qquad Update denoising model $\epsilon_{\theta}$ with complete data by Equation \ref{eq:l_dm} \\
    {\bf end for} \\
    
    {\bf Step 3:} Train all MLP layers of conntrastive clustering \\
    \qquad Impute the missing views by diffusion completion \\
    {\bf for} $t=0$ to $T_{dm}-1$ {\bf do} \\
    \qquad Update all MLP layers with complete data and imputation data by \\
    \qquad Equation \ref{eq:l_clu} \\
    {\bf end for} \\

    Compute cluster predction $\bar{Y}$ by Equation \ref{eq:prediction}\\
    \bottomrule
  \end{tabular}
\end{table}

\subsection{Contrastive Clustering}

With the remarkable achievements of contrastive learning in the field of clustering \cite{li2021contrastive}, in recent years, more and more multi-view clustering studies have used contrastive learning to achieve clustering \cite{lin2022dual, tang2022deep, Xu_2022_CVPR}. By completing the missing views through diffusion models, we can obtain the latent representations for all the views of the sample. The latent representation of each view is fed into the view-specific MLP layers and the view-sharing MLP layers, and thus the latent representation of each view is projected into the shared space. Finally, the high-level semantic information and the class information are obtained through contrastive learning separately. We adopted the spectral contrastive loss function \cite{haochen2021provable} to maximize the similarity of high-level semantic information between the same sample views and minimize the similarity between different sample views:
\begin{equation}
    L_{H} = -\frac{2}{n}\sum_{i=1}^n(H_i^1)^T+\frac{1}{n(n-1)}\sum_{i=1}^n\sum_{j\neq i}((H_i^1)^TH_i^2)^2
\end{equation}

Different views of multi-view data contain the same category information, which can be obtained by contrastive learning:

\begin{equation}
\begin{split}
    L_{C} = &-\frac{1}{k}\sum_{i=1}^k \left[ 
    \log \frac{e^{({\widehat{Y}_i^1})^T\widehat{Y}_i^2}}{\sum_{j\neq i} e^{(\widehat{Y}_i^1)^T\widehat{Y}_j^1}}
    + log\frac{e^{({\widehat{Y}_i^1})^T\widehat{Y}_i^2}}{\sum_{j\neq i}e^{(\widehat{Y}_i^2)^T\widehat{Y}_j^2}}
    \right] \\
    &+ \sum_{m=1}^2 \sum_{j=1}^k \widehat{Y}_j^m\log(\widehat{Y}_j^m)
    \end{split}
\end{equation}

Here $k$ is the number of categories, $\widehat{Y}_j^m$ is the prediction probability of category $j$ from the $v$-th view, and the second term is a constraint term \cite{van2020scan}, which is used to prevent samples from being classified into the same category. Finally, our clustering objective is:
\begin{equation}
    L_{clu} = L_H+L_C
    \label{eq:l_clu}
\end{equation}

Finally, we perform a simple addition operation on the clustering results from multiple views to obtain the final prediction:
\begin{equation}
    \bar{Y} = \arg \max (\widehat{Y}^1 + \widehat{Y}^2)
    \label{eq:prediction}
\end{equation}

The final objective function of our proposed Incomplete Multi-view Clustering via Diffusion Completion (IMVCDC) is:
\begin{equation}
    L_{IMVCDC} = L_{rec} + L_{dm} + L_{clu}
    \label{eq:final_objective}
\end{equation}

\subsection{Implementation Details}
As shown in Figure \ref{fig:overview}, IMVCDC consists of three modules, namely autoencoders, diffusion completion module, and contrastive clustering module. Our autoencoders are based on existing research \cite{esser2021taming} and simply use reconstruction loss to train the autoencoders. The diffusion model framework of \cite{rombach2022high} is used as our diffusion completion framework. Since the autoencoder can extract high-quality latent representations and the diffusion completion module can robustly and effectively complete missing views, we only use simple MLP layers to complete the contrastive clustering module.

In the training phase, AdamW\cite{loshchilov2017decoupled} gradient update algorithm was used to update all modules. We adopt a step-by-step optimization strategy to optimize the parameters of the three modules. Firstly, we used Equation \ref{eq:l_rec} to update and optimize the autoencoder so that the autoencoder could obtain low-dimensional and high-quality latent representation. Then, the diffusion completion model was optimized by Equation \ref{eq:l_dm} to complete the missing views data and obtain the latent representation of all views of all samples. Finally, Equation \ref{eq:l_clu} was used to optimize the contrastive clustering module, and the final prediction result was obtained through Equation \ref{eq:prediction}. 

\section{Experiments}
In this section, we compare IMVCDC with state-of-the-art incomplete multi-view clustering methods on four multi-view datasets.

\begin{table*}
\centering
  \begin{tabular}{lcccccccccccc}
    \toprule
    \multirow{2}{*}{Method} & \multicolumn{3}{c}{Multi-Fashion} & \multicolumn{3}{c}{Multi-Coil20} & \multicolumn{3}{c}{NoisyMNIST} & \multicolumn{3}{c}{NoisyMNIST-Product} \\
    & ACC & NMI & ARI & ACC & NMI & ARI & ACC & NMI & ARI & ACC & NMI & ARI \\
    \midrule
    CDIMC-Net & 46.91 & 60.78 & 30.34 & 77.64 & 84.97 & 71.55 & 40.77 & 40.37 & 25.56 & 58.40 & 53.81 & 34.57 \\
    COMPLETER & 77.95 & 76.70 & 67.14 & 77.64 & 88.08 & 73.74 & 78.52 & 72.00 & 66.45 & 71.55 & 63.59 & 54.68 \\
    OS-FL-IMVC & 49.06 & 45.93 & 32.70 & 52.95 & 64.39 & 43.02 & - & - & - & 53.98 & 45.10 & 34.56 \\
    IMVTSC-MVI & 68.49 & 67.17 & 57.32 & 83.39 & \underline{91.29} & \underline{79.82} & - & - & - & 42.85 & 41.95 & 22.43 \\
    DIMVC & 63.35 & 68.81 & 53.95 & 69.87 & 77.79 & 62.49 & 64.37 & 65.48 & 52.90 & \underline{83.60} & \underline{70.91} & \underline{68.68} \\
    DCP & 79.37 & \textbf{78.73} & 70.67 & 76.53 & 89.33 & 63.26 & 83.52 & 75.33 & 69.53 & 75.85 & 70.90 & 64.96 \\
    DSIMVC & \underline{83.07} & 77.55 & \underline{71.88} & \underline{86.53} & 89.86 & 79.81 & \underline{86.56} & \underline{76.06} & \underline{74.14} & 75.96 & 66.06 & 61.16 \\
    IMVCDC & \textbf{86.48} & \underline{77.93} & \textbf{74.46} & \textbf{98.26} & \textbf{97.35} & \textbf{96.42} & \textbf{92.61} & \textbf{82.51} & \textbf{84.43} & \textbf{87.26} & \textbf{75.08} & \textbf{74.56} \\
    \bottomrule
  
  \end{tabular}
  \caption{Comparison of clustering performance on four datasets. "-" indicates that the result is unknown due to out of memory. The first and second best results are shown in bold and sliding lines, respectively.}
  
  \label{tab:experiments}
\end{table*}

\subsection{Experimental Setup}
We use four multi-view datasets for our experiments. We constructed the Multi-Fashion two-view dataset from Fashion \cite{xiao2017fashion} by randomly selecting two products from the same category as the two-view data and constructing 10k samples in total. The Multi-Coil20 dataset selects different angles of the same object from Coil20 \cite{nene1996columbia} to construct a two-view dataset with 1440 samples in total. NoisyMNIST \cite{wang2015deep} takes the original MNIST \cite{lecun1998gradient} handwritten data graph as the first view and randomly selects handwritten digits of the same class with white Gaussian noise as the second view. Since some methods cannot handle the data size of 70k, we select a subset of 20k for our experiments.  In addition, we construct the NoisyDigit-Product dataset. The first view is hand-written digital data with white Gaussian noise from NoisyMNIST, and the second view is Fashion. There is a one-to-one correspondence between numbers and product categories with a total of 10k samples.

We compare IMVCDC with the following IMVC methods: CDIMC-net \cite{ijcai2020p447}, COMPLETER \cite{lin2021completer}, OS-FL-IMVC \cite{zhang2021one}, IMVTSC-MVI \cite{wen2021unified}, DIMVC \cite{xu2022deep}, DCP \cite{lin2022dual}, DSIMVC \cite{tang2022deep}. The missing rate of incomplete multi-view data is denoted as $\eta = \widehat{n}/n$, where $\widehat{n}$ denotes the number of incomplete multi-view samples, and $n$ is the number of the entire dataset. We tested all methods with the missing rate n=0.5. Three evaluation indexes were used: Accuracy (ACC), Normalized Mutual Information (NMI), and Adjusted Rand Index (ARI). Higher values of these metrics indicate better clustering performance.

\begin{figure}
  \centering
  \subfigure[10 epoch]{
  \label{fig:t-sne-a}
  \includegraphics[width=0.2\linewidth]{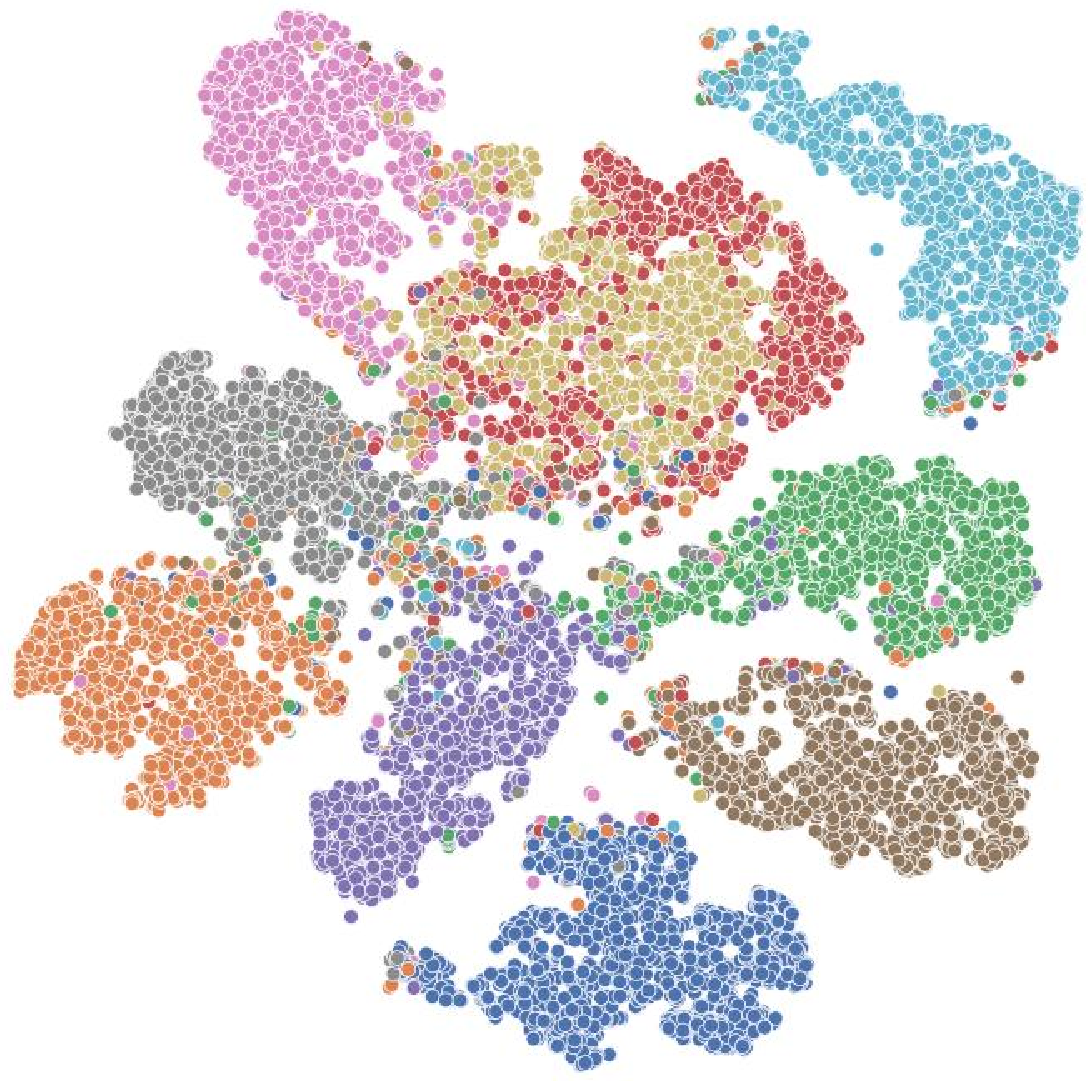}
  }
  \hfill
  \subfigure[50 epoch]{
  \label{fig:t-sne-b}
  \includegraphics[width=0.2\linewidth]{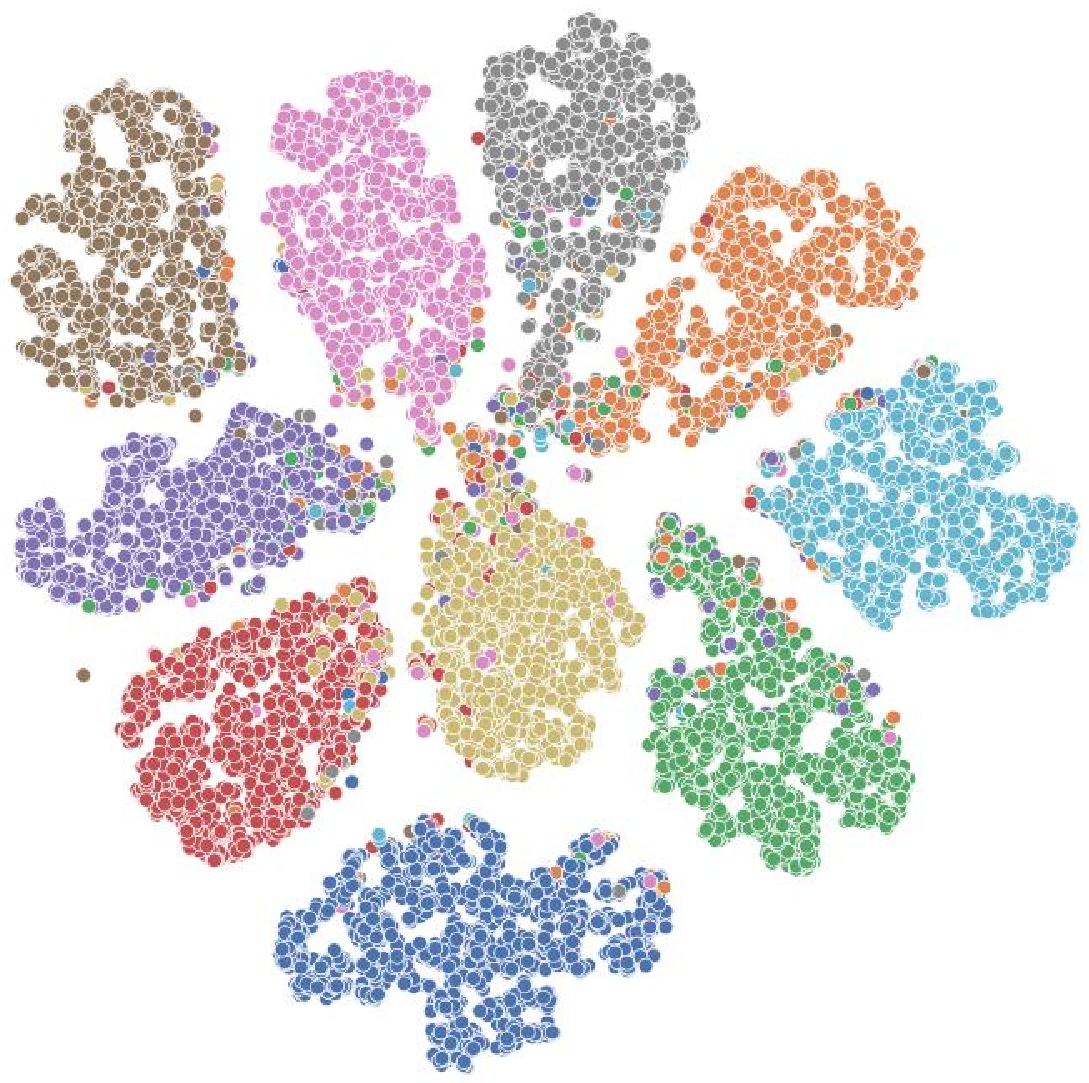}
  }
  \hfill
  \subfigure[100 epoch]{
  \label{fig:t-sne-c}
  \includegraphics[width=0.2\linewidth]{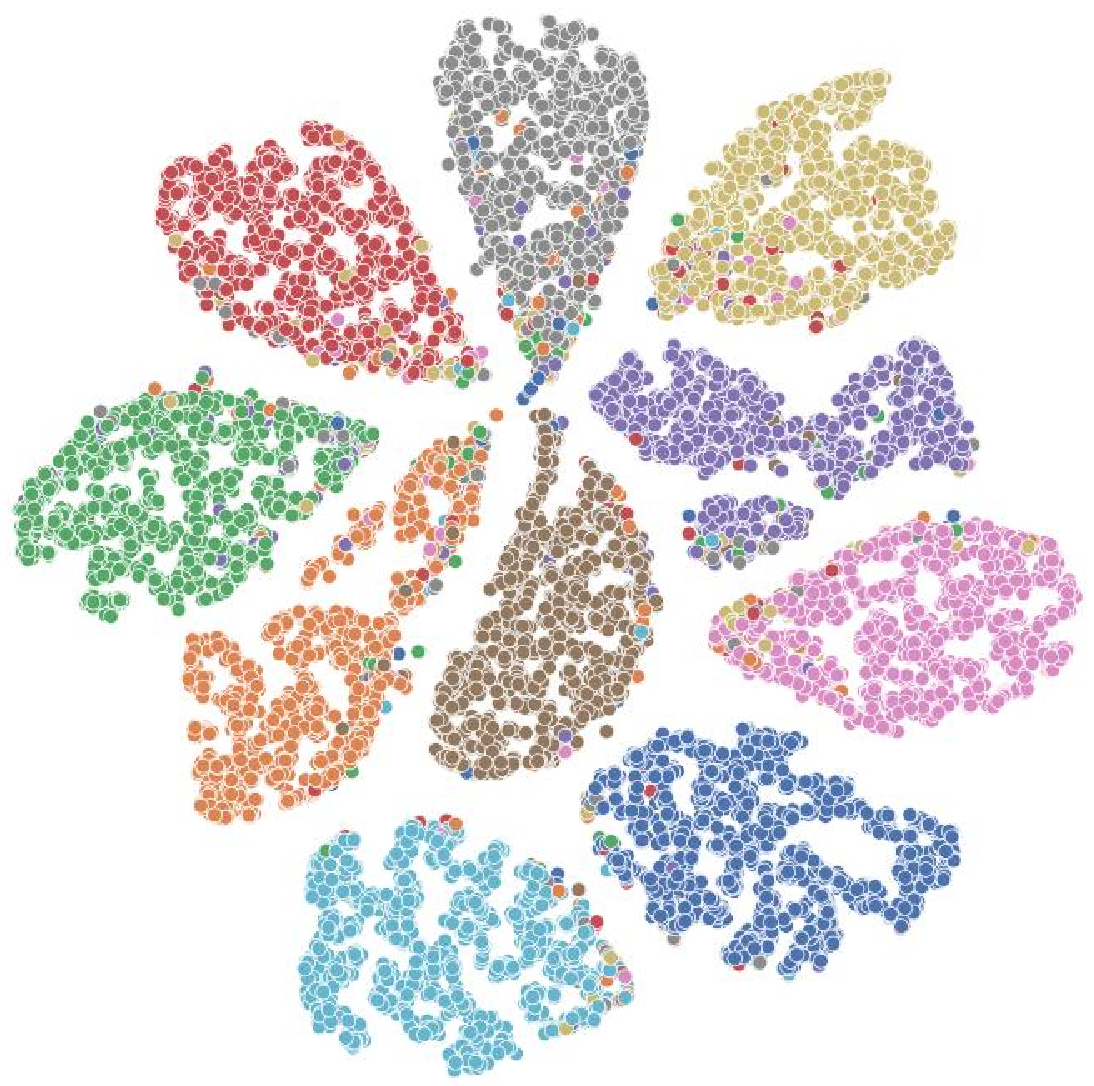}
  }
  \hfill
  \subfigure[200 epoch]{
  \label{fig:t-sne-d}
  \includegraphics[width=0.2\linewidth]{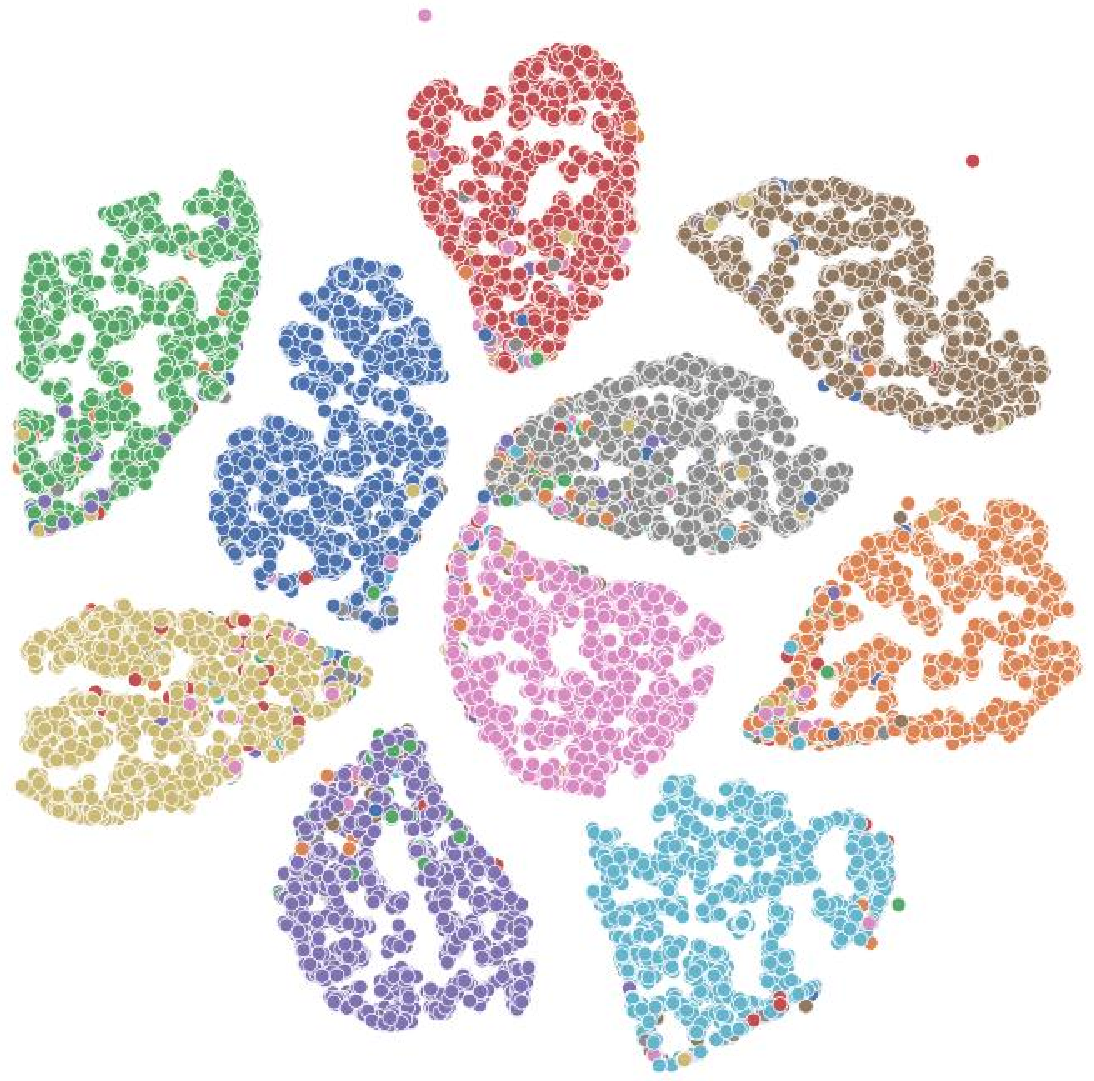}
  }
  \caption{Visualization of t-sne on NoisyMNIST with increasing training iterations for contrastive clustering..}
  \label{fig:t-sne}
\end{figure}

\begin{figure}[t]
  \centering
   \includegraphics[width=0.8\linewidth]{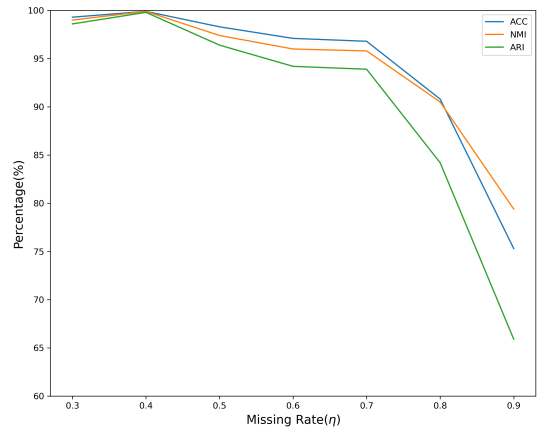}
   \caption{Performance with different missing rates on Multi-Coil20}
   \label{fig:missing_rate}
\end{figure}

\begin{figure}[t]
  \centering
   \includegraphics[width=0.6\linewidth]{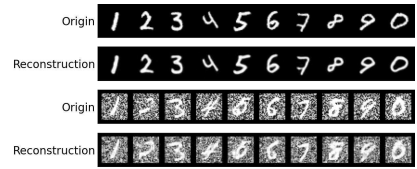}
   \caption{View reconstruction is performed on NoisyMNIST. Line 1 and 3 are the original views, and Line 2 and 4 are the reconstructions of the original views.}
   \label{fig:view_rec}
\end{figure}

\begin{figure}[t]
   \includegraphics[width=0.8\linewidth]{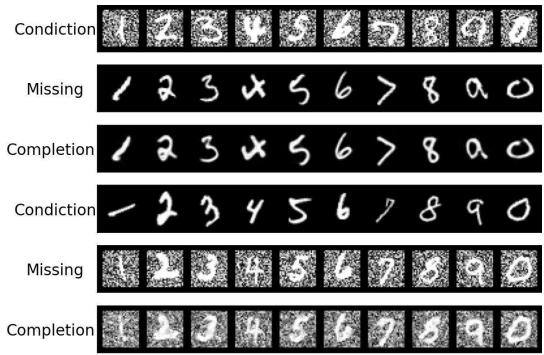}
   \caption{Visualization of diffusion completion on NoisyMNIST. Line 1 and 4 are the conditional views of the observables, Line 2 and 5 are the missing views, and Line 3 and 6 are the recovered results using diffusion completion.}
   \label{fig:view_dm}
\end{figure}

\subsection{Experimental Results}
As shown in Table \ref{tab:experiments}, IMVCDC outperforms these state-of-the-art methods on all four datasets. In terms of ACC, IMVCDC outperforms the optimal performance of these state-of-the-art methods in four datasets by a margin of 3.41\% on Multi-Fashion, 11.73\% on Multi-Coil20, 6.05\% on NoisyMNIST, and 3.66\% on NoisyDigit-Product. For ARI, IMVCDC significantly outperforms all methods. IMVCDC surpasses the best method by 12.92\% on Multi-Coil20, 6.45\% on NoisyMNIST, and 4.17\% on NoisyDigit-Product in terms of NMI. While the NMI of IMVCDC in Multi-Fashion is slightly below DCP by 0.90\%, the performance of ACC and ARI outperform DCP by 7.11\% and 3.79\%, respectively. The overall performance is better than that of DCP. Experimental results prove the superiority of our proposed IMVCDC over other methods, which is due to IMVCDC's concise and efficient framework, namely, the latent representation of view is obtained through view reconstruction, then the diffusion model is used to complete the missing views, and finally the category information is obtained by contrastive learning.

\subsection{Performance with Different Missing Rates}
To further validate the effectiveness of our method, we perform experiments on Multi-Coil20 with missing rate intervals of 0.1 from 0.3 to 0.9. As shown in Figure \ref{fig:missing_rate}, when the missing rate is above 0.7, the performance of all three metrics is above 90\%. When the missing rate is 0.8, ACC and NMI are around 90\%, and ARI is around 85\%. The performance of IMVCDC significantly degrades only when there is a huge missing rate of 0.9. However, the performance of all three metrics is above 60\%, which indicates the effectiveness of our IMVCDR method.

\subsection{Visualization}

In this section, the theoretical results of our proposed method are experimentally verified. Experiments are conducted on the NoisyMNIST dataset to visualize view reconstruction, diffusion completion, and common representation. In the experiments, the missing rate is fixed at 0.5.

Since we use a step-by-step optimization strategy to optimize the IMVCDC framework, view reconstruction is the basis of our framework. As shown in Figure \ref{fig:view_rec}, the autoencoder reconstructs the original view well, which verifies that the view reconstruction module can extract low-dimensional and efficient latent representations. Unlike most incomplete multi-view methods, IMVCDC uses diffusion models to complete the missing views in the latent space. To demonstrate the recoverability of our proposed IMVCDC, Figure \ref{fig:view_dm} visualizes the completion results for some samples from NoisyMNIST. Line 1 and 4 are the observable views as conditions for diffusion completion. Line 2 and 5 are the missing views, while line 3 and 6 are the recovered views using diffusion completion. In short, IMVCDC recovers important information from missing views and preserves the original data distribution features of the views.

In addition to the visualizations described above, we also performed t-sne \cite{van2008visualizing} visualizations of common representations learned from comparative clustering. As shown in Figure \ref{fig:t-sne}, the clustering performance becomes more and more compact and separated as the number of iterations increases, which indicates that our proposed IMVCDC framework can capture the semantic information of multi-view data well.

\subsection{Ablation Studies}

\begin{table}
  \centering
  \begin{tabular}{lccc}
    \toprule
    & ACC & NMI & ARI \\
    \midrule
    $L_{rec}$ & 34.03 & 50.35 & 22.69 \\
    $L_{rec}+L_{dm}$ & 74.49 & 82.13 & 70.01 \\
    $L_{rec}+L_{clu}$ & 74.03 & 73.99 & 33.77 \\
    $L_{rec}+L_{dm}+L_{clu}$ & 98.26 & 97.35 & 96.42 \\
    \bottomrule
  \end{tabular}
  \caption{Ablation experiments are performed on Multi-Coil20 dataset with a miss rate of 0.5}
  \label{tab:ablation}
\end{table}

Ablation experiments are conducted to demonstrate the effectiveness of view reconstruction, diffusion completion, and contrastive clustering. Since our optimization strategy is a step-by-step optimization strategy, we compute our clustering performance after each step. After the first view reconstruction and the second diffusion completion step, we simply compute the clustering performance using the K-means method. In the absence of the diffusion completion module, we will simply use the zero-padding method to complete the missing views. The experimental results are shown in Table \ref{tab:ablation}, as the number of modules increases, the clustering performance becomes better and better, indicating that the design of each module has done its part. We observe that the clustering performance with missing diffusion completion performs better than with missing contrastive clustering module. This also reflects the fact that the imputed method is better than the imputed-free method because it completes the missing views and fully exploits the complementary information between the views.

\section{Conclusion}
In this paper, we propose a straightforward and powerful IMVCDC framework. Based on the latent space of views learned by the auto-encoder, Incomplete Multi-view Clustering is implemented by using diffusion completion to recover missing views and contrastive learning to extract consistent information from multi-view data. In short, incomplete multi-view clustering is divided into three steps: 1) Feature extraction of each view; 2) Impute the missing views; 3) Multi-view clustering. Such a simple and unified framework would provide the community with novel insights on how to mitigate the effects of missing views and multi-view clustering. In the future, we plan to further explore our theoretical framework in three views and above for multi-view learning and applications in multiple modalities.


%





\ifCLASSOPTIONcaptionsoff
  \newpage
\fi

\bibliographystyle{IEEEtran}
\bibliography{cite}

\begin{thebibliography}{10}
\providecommand{\url}[1]{#1}
\csname url@samestyle\endcsname
\providecommand{\newblock}{\relax}
\providecommand{\bibinfo}[2]{#2}
\providecommand{\BIBentrySTDinterwordspacing}{\spaceskip=0pt\relax}
\providecommand{\BIBentryALTinterwordstretchfactor}{4}
\providecommand{\BIBentryALTinterwordspacing}{\spaceskip=\fontdimen2\font plus
\BIBentryALTinterwordstretchfactor\fontdimen3\font minus
  \fontdimen4\font\relax}
\providecommand{\BIBforeignlanguage}[2]{{%
\expandafter\ifx\csname l@#1\endcsname\relax
\typeout{** WARNING: IEEEtran.bst: No hyphenation pattern has been}%
\typeout{** loaded for the language `#1'. Using the pattern for}%
\typeout{** the default language instead.}%
\else
\language=\csname l@#1\endcsname
\fi
#2}}
\providecommand{\BIBdecl}{\relax}
\BIBdecl

\bibitem{Xu_2021_ICCV}
J.~Xu, Y.~Ren, H.~Tang, X.~Pu, X.~Zhu, M.~Zeng, and L.~He, ``Multi-vae:
  Learning disentangled view-common and view-peculiar visual representations
  for multi-view clustering,'' in \emph{Proceedings of the IEEE/CVF
  International Conference on Computer Vision (ICCV)}, October 2021, pp.
  9234--9243.

\bibitem{NEURIPS2021_10c66082}
E.~Pan and Z.~Kang, ``Multi-view contrastive graph clustering,'' in
  \emph{Advances in Neural Information Processing Systems}, M.~Ranzato,
  A.~Beygelzimer, Y.~Dauphin, P.~Liang, and J.~W. Vaughan, Eds., vol.~34.\hskip
  1em plus 0.5em minus 0.4em\relax Curran Associates, Inc., 2021, pp.
  2148--2159.

\bibitem{Xu_2022_CVPR}
J.~Xu, H.~Tang, Y.~Ren, L.~Peng, X.~Zhu, and L.~He, ``Multi-level feature
  learning for contrastive multi-view clustering,'' in \emph{Proceedings of the
  IEEE/CVF Conference on Computer Vision and Pattern Recognition (CVPR)}, June
  2022, pp. 16\,051--16\,060.

\bibitem{wang2019gmc}
H.~Wang, Y.~Yang, and B.~Liu, ``Gmc: Graph-based multi-view clustering,''
  \emph{IEEE Transactions on Knowledge and Data Engineering}, vol.~32, no.~6,
  pp. 1116--1129, 2019.

\bibitem{liu2021one}
X.~Liu, L.~Liu, Q.~Liao, S.~Wang, Y.~Zhang, W.~Tu, C.~Tang, J.~Liu, and E.~Zhu,
  ``One pass late fusion multi-view clustering,'' in \emph{International
  Conference on Machine Learning}.\hskip 1em plus 0.5em minus 0.4em\relax PMLR,
  2021, pp. 6850--6859.

\bibitem{peng2019comic}
X.~Peng, Z.~Huang, J.~Lv, H.~Zhu, and J.~T. Zhou, ``Comic: Multi-view
  clustering without parameter selection,'' in \emph{International conference
  on machine learning}.\hskip 1em plus 0.5em minus 0.4em\relax PMLR, 2019, pp.
  5092--5101.

\bibitem{chen2020multi}
M.-S. Chen, L.~Huang, C.-D. Wang, and D.~Huang, ``Multi-view clustering in
  latent embedding space,'' in \emph{Proceedings of the AAAI conference on
  artificial intelligence}, vol.~34, no.~04, 2020, pp. 3513--3520.

\bibitem{lin2021completer}
Y.~Lin, Y.~Gou, Z.~Liu, B.~Li, J.~Lv, and X.~Peng, ``Completer: Incomplete
  multi-view clustering via contrastive prediction,'' in \emph{Proceedings of
  the IEEE/CVF conference on computer vision and pattern recognition}, 2021,
  pp. 11\,174--11\,183.

\bibitem{lin2022dual}
Y.~Lin, Y.~Gou, X.~Liu, J.~Bai, J.~Lv, and X.~Peng, ``Dual contrastive
  prediction for incomplete multi-view representation learning,'' \emph{IEEE
  Transactions on Pattern Analysis and Machine Intelligence}, 2022.

\bibitem{ijcai2020p447}
\BIBentryALTinterwordspacing
J.~Wen, Z.~Zhang, Y.~Xu, B.~Zhang, L.~Fei, and G.-S. Xie, ``Cdimc-net:
  Cognitive deep incomplete multi-view clustering network,'' in
  \emph{Proceedings of the Twenty-Ninth International Joint Conference on
  Artificial Intelligence, {IJCAI-20}}, C.~Bessiere, Ed.\hskip 1em plus 0.5em
  minus 0.4em\relax International Joint Conferences on Artificial Intelligence
  Organization, 7 2020, pp. 3230--3236, main track. [Online]. Available:
  \url{https://doi.org/10.24963/ijcai.2020/447}
\BIBentrySTDinterwordspacing

\bibitem{zhang2021one}
Y.~Zhang, X.~Liu, S.~Wang, J.~Liu, S.~Dai, and E.~Zhu, ``One-stage incomplete
  multi-view clustering via late fusion,'' in \emph{Proceedings of the 29th ACM
  International Conference on Multimedia}, 2021, pp. 2717--2725.

\bibitem{wen2021unified}
J.~Wen, Z.~Zhang, Z.~Zhang, L.~Zhu, L.~Fei, B.~Zhang, and Y.~Xu, ``Unified
  tensor framework for incomplete multi-view clustering and missing-view
  inferring,'' in \emph{Proceedings of the AAAI conference on artificial
  intelligence}, vol.~35, no.~11, 2021, pp. 10\,273--10\,281.

\bibitem{xu2022deep}
J.~Xu, C.~Li, Y.~Ren, L.~Peng, Y.~Mo, X.~Shi, and X.~Zhu, ``Deep incomplete
  multi-view clustering via mining cluster complementarity,'' in
  \emph{Proceedings of the AAAI Conference on Artificial Intelligence},
  vol.~36, no.~8, 2022, pp. 8761--8769.

\bibitem{tang2022deep}
H.~Tang and Y.~Liu, ``Deep safe incomplete multi-view clustering: Theorem and
  algorithm,'' in \emph{International Conference on Machine Learning}.\hskip
  1em plus 0.5em minus 0.4em\relax PMLR, 2022, pp. 21\,090--21\,110.

\bibitem{wang2021generative}
Q.~Wang, Z.~Ding, Z.~Tao, Q.~Gao, and Y.~Fu, ``Generative partial multi-view
  clustering with adaptive fusion and cycle consistency,'' \emph{IEEE
  Transactions on Image Processing}, vol.~30, pp. 1771--1783, 2021.

\bibitem{li2014partial}
S.-Y. Li, Y.~Jiang, and Z.-H. Zhou, ``Partial multi-view clustering,'' in
  \emph{Proceedings of the AAAI conference on artificial intelligence},
  vol.~28, no.~1, 2014.

\bibitem{wen2021structural}
J.~Wen, Z.~Wu, Z.~Zhang, L.~Fei, B.~Zhang, and Y.~Xu, ``Structural deep
  incomplete multi-view clustering network,'' in \emph{Proceedings of the 30th
  ACM International Conference on Information \& Knowledge Management}, 2021,
  pp. 3538--3542.

\bibitem{xu2019adversarial}
C.~Xu, Z.~Guan, W.~Zhao, H.~Wu, Y.~Niu, and B.~Ling, ``Adversarial incomplete
  multi-view clustering.'' in \emph{IJCAI}, vol.~7, 2019, pp. 3933--3939.

\bibitem{zhao2016incomplete}
H.~Zhao, H.~Liu, and Y.~Fu, ``Incomplete multi-modal visual data grouping.'' in
  \emph{IJCAI}, 2016, pp. 2392--2398.

\bibitem{wang2022adversarial}
Q.~Wang, Z.~Tao, W.~Xia, Q.~Gao, X.~Cao, and L.~Jiao, ``Adversarial multiview
  clustering networks with adaptive fusion,'' \emph{IEEE transactions on neural
  networks and learning systems}, 2022.

\bibitem{sohl2015deep}
J.~Sohl-Dickstein, E.~Weiss, N.~Maheswaranathan, and S.~Ganguli, ``Deep
  unsupervised learning using nonequilibrium thermodynamics,'' in
  \emph{International Conference on Machine Learning}.\hskip 1em plus 0.5em
  minus 0.4em\relax PMLR, 2015, pp. 2256--2265.

\bibitem{ho2020denoising}
J.~Ho, A.~Jain, and P.~Abbeel, ``Denoising diffusion probabilistic models,''
  \emph{Advances in Neural Information Processing Systems}, vol.~33, pp.
  6840--6851, 2020.

\bibitem{song2019generative}
Y.~Song and S.~Ermon, ``Generative modeling by estimating gradients of the data
  distribution,'' \emph{Advances in neural information processing systems},
  vol.~32, 2019.

\bibitem{rombach2022high}
R.~Rombach, A.~Blattmann, D.~Lorenz, P.~Esser, and B.~Ommer, ``High-resolution
  image synthesis with latent diffusion models,'' in \emph{Proceedings of the
  IEEE/CVF Conference on Computer Vision and Pattern Recognition}, 2022, pp.
  10\,684--10\,695.

\bibitem{dhariwal2021diffusion}
P.~Dhariwal and A.~Nichol, ``Diffusion models beat gans on image synthesis,''
  \emph{Advances in Neural Information Processing Systems}, vol.~34, pp.
  8780--8794, 2021.

\bibitem{kong2020diffwave}
Z.~Kong, W.~Ping, J.~Huang, K.~Zhao, and B.~Catanzaro, ``Diffwave: A versatile
  diffusion model for audio synthesis,'' \emph{arXiv preprint
  arXiv:2009.09761}, 2020.

\bibitem{chen2020wavegrad}
N.~Chen, Y.~Zhang, H.~Zen, R.~J. Weiss, M.~Norouzi, and W.~Chan, ``Wavegrad:
  Estimating gradients for waveform generation,'' \emph{arXiv preprint
  arXiv:2009.00713}, 2020.

\bibitem{ramesh2022hierarchical}
A.~Ramesh, P.~Dhariwal, A.~Nichol, C.~Chu, and M.~Chen, ``Hierarchical
  text-conditional image generation with clip latents,'' \emph{arXiv preprint
  arXiv:2204.06125}, 2022.

\bibitem{saharia2022photorealistic}
C.~Saharia, W.~Chan, S.~Saxena, L.~Li, J.~Whang, E.~Denton, S.~K.~S.
  Ghasemipour, B.~K. Ayan, S.~S. Mahdavi, R.~G. Lopes \emph{et~al.},
  ``Photorealistic text-to-image diffusion models with deep language
  understanding,'' \emph{arXiv preprint arXiv:2205.11487}, 2022.

\bibitem{li2022srdiff}
H.~Li, Y.~Yang, M.~Chang, S.~Chen, H.~Feng, Z.~Xu, Q.~Li, and Y.~Chen,
  ``Srdiff: Single image super-resolution with diffusion probabilistic
  models,'' \emph{Neurocomputing}, vol. 479, pp. 47--59, 2022.

\bibitem{saharia2022image}
C.~Saharia, J.~Ho, W.~Chan, T.~Salimans, D.~J. Fleet, and M.~Norouzi, ``Image
  super-resolution via iterative refinement,'' \emph{IEEE Transactions on
  Pattern Analysis and Machine Intelligence}, 2022.

\bibitem{brock2018large}
A.~Brock, J.~Donahue, and K.~Simonyan, ``Large scale gan training for high
  fidelity natural image synthesis,'' \emph{arXiv preprint arXiv:1809.11096},
  2018.

\bibitem{liu2019spectral}
K.~Liu, W.~Tang, F.~Zhou, and G.~Qiu, ``Spectral regularization for combating
  mode collapse in gans,'' in \emph{Proceedings of the IEEE/CVF international
  conference on computer vision}, 2019, pp. 6382--6390.

\bibitem{ho2022classifier}
J.~Ho and T.~Salimans, ``Classifier-free diffusion guidance,'' \emph{arXiv
  preprint arXiv:2207.12598}, 2022.

\bibitem{nichol2021glide}
A.~Nichol, P.~Dhariwal, A.~Ramesh, P.~Shyam, P.~Mishkin, B.~McGrew,
  I.~Sutskever, and M.~Chen, ``Glide: Towards photorealistic image generation
  and editing with text-guided diffusion models,'' \emph{arXiv preprint
  arXiv:2112.10741}, 2021.

\bibitem{saharia2022palette}
C.~Saharia, W.~Chan, H.~Chang, C.~Lee, J.~Ho, T.~Salimans, D.~Fleet, and
  M.~Norouzi, ``Palette: Image-to-image diffusion models,'' in \emph{ACM
  SIGGRAPH 2022 Conference Proceedings}, 2022, pp. 1--10.

\bibitem{pinaya2022brain}
W.~H. Pinaya, P.-D. Tudosiu, J.~Dafflon, P.~F. Da~Costa, V.~Fernandez,
  P.~Nachev, S.~Ourselin, and M.~J. Cardoso, ``Brain imaging generation with
  latent diffusion models,'' in \emph{Deep Generative Models: Second MICCAI
  Workshop, DGM4MICCAI 2022, Held in Conjunction with MICCAI 2022, Singapore,
  September 22, 2022, Proceedings}.\hskip 1em plus 0.5em minus 0.4em\relax
  Springer, 2022, pp. 117--126.

\bibitem{vaswani2017attention}
A.~Vaswani, N.~Shazeer, N.~Parmar, J.~Uszkoreit, L.~Jones, A.~N. Gomez,
  {\L}.~Kaiser, and I.~Polosukhin, ``Attention is all you need,''
  \emph{Advances in neural information processing systems}, vol.~30, 2017.

\bibitem{li2021contrastive}
Y.~Li, P.~Hu, Z.~Liu, D.~Peng, J.~T. Zhou, and X.~Peng, ``Contrastive
  clustering,'' in \emph{Proceedings of the AAAI Conference on Artificial
  Intelligence}, vol.~35, no.~10, 2021, pp. 8547--8555.

\bibitem{haochen2021provable}
J.~Z. HaoChen, C.~Wei, A.~Gaidon, and T.~Ma, ``Provable guarantees for
  self-supervised deep learning with spectral contrastive loss,''
  \emph{Advances in Neural Information Processing Systems}, vol.~34, pp.
  5000--5011, 2021.

\bibitem{van2020scan}
W.~Van~Gansbeke, S.~Vandenhende, S.~Georgoulis, M.~Proesmans, and L.~Van~Gool,
  ``Scan: Learning to classify images without labels,'' in \emph{Computer
  Vision--ECCV 2020: 16th European Conference, Glasgow, UK, August 23--28,
  2020, Proceedings, Part X}.\hskip 1em plus 0.5em minus 0.4em\relax Springer,
  2020, pp. 268--285.

\bibitem{esser2021taming}
P.~Esser, R.~Rombach, and B.~Ommer, ``Taming transformers for high-resolution
  image synthesis,'' in \emph{Proceedings of the IEEE/CVF conference on
  computer vision and pattern recognition}, 2021, pp. 12\,873--12\,883.

\bibitem{loshchilov2017decoupled}
I.~Loshchilov and F.~Hutter, ``Decoupled weight decay regularization,''
  \emph{arXiv preprint arXiv:1711.05101}, 2017.

\bibitem{xiao2017fashion}
H.~Xiao, K.~Rasul, and R.~Vollgraf, ``Fashion-mnist: a novel image dataset for
  benchmarking machine learning algorithms,'' \emph{arXiv preprint
  arXiv:1708.07747}, 2017.

\bibitem{nene1996columbia}
S.~A. Nene, S.~K. Nayar, H.~Murase \emph{et~al.}, ``Columbia object image
  library (coil-20),'' 1996.

\bibitem{wang2015deep}
W.~Wang, R.~Arora, K.~Livescu, and J.~Bilmes, ``On deep multi-view
  representation learning,'' in \emph{International conference on machine
  learning}.\hskip 1em plus 0.5em minus 0.4em\relax PMLR, 2015, pp. 1083--1092.

\bibitem{lecun1998gradient}
Y.~LeCun, L.~Bottou, Y.~Bengio, and P.~Haffner, ``Gradient-based learning
  applied to document recognition,'' \emph{Proceedings of the IEEE}, vol.~86,
  no.~11, pp. 2278--2324, 1998.

\bibitem{van2008visualizing}
L.~Van~der Maaten and G.~Hinton, ``Visualizing data using t-sne.''
  \emph{Journal of machine learning research}, vol.~9, no.~11, 2008.

\end{thebibliography}




\end{document}